*Article*

# Automatic Crack Detection on Road Pavements Using Encoder-Decoder Architecture

**Zhun Fan** [1,2], **Chong Li** [1,2,3], **Ying Chen** [1,2], **Jiahong Wei** [1,2], **Giuseppe Loprencipe** [3,*], **Xiaopeng Chen** [4] **and Paola Di Mascio** [3]

[1] Key Lab of Digital Signal and Image Processing of Guangdong Province, Department of Electronic and information Engineering, College of Engineering, Shantou University, Shan'tou, 515063, China; zfan@stu.edu.cn (Z.F.); chongli1217@163.com (C.L.); 19ychen1@stu.edu.cn (Y.C.); 19jhwei@stu.edu.cn (J.W.)

[2] Department of Electronic and information Engineering, College of Engineering, Shantou University, Shan'tou, 515063, China

[3] Department of Civil, Constructional and Environmental Engineering, Sapienza University of Rome, 00184 Rome, Italy; paola.dimascio@uniroma1.it (P.D.M.)

[4] Department of Industrial Engineering, Pusan National University, Busan 609735, Korea; xiaopengchen388@gmail.com (X.C.)

\* Correspondence: giuseppe.loprencipe@uniroma1.it (G.L.)

**Abstract:** Automatic crack detection from images is an important task that is adopted to ensure road safety and durability for Portland cement concrete (PCC) and asphalt concrete (AC) pavement. Pavement failure depends on a number of causes including water intrusion, stress from heavy loads, and all the climate effects. Generally, cracks are the first distress that arises on road surfaces and proper monitoring and maintenance to prevent cracks from spreading or forming is important. Conventional algorithms to identify cracks on road pavements are extremely time-consuming and high cost. Many cracks show complicated topological structures, oil stains, poor continuity, and low contrast, which are difficult for defining crack features. Therefore, the automated crack detection algorithm is a key tool to improve the results. Inspired by the development of deep learning in computer vision and object detection, the proposed algorithm considers an encoder-decoder architecture with hierarchical feature learning and dilated convolution, named U-Hierarchical Dilated Network (U-HDN), to perform crack detection in an end-to-end method. Crack characteristics with multiple context information are automatically able to learn and perform end-to-end crack detection. Then, a multi-dilation module embedded in an encoder-decoder architecture is proposed. The crack features of multiple context sizes can be integrated into the multi-dilation module by dilation convolution with different dilatation rates, which can obtain much more cracks information. Finally, the hierarchical feature learning module is designed to obtain a multi-scale features from the high to low- level convolutional layers, which are integrated to predict pixel-wise crack detection. Some experiments on public crack databases using 118 images were performed and the results were compared with those obtained with other methods on the same images. The results show that the proposed U-HDN method achieves high performance because it can extract and fuse different context sizes and different levels of feature maps than other algorithms.

## 1. Introduction

*1.1. Motivation*

Cracks are common distresses in both concrete and asphalt pavements. Different types of cracks can be observed due to different causes: road surface aging, climate, and traffic load. The methods currently used for road and airport pavement management system (PMS) [1,2] generally used for the classification of cracks provided by Shahin [3] and adopted by the international standard American Society for Testing and Materials ( ASTM ) [4]. The classification is defined on crack characteristic and causes as listed in Table 1 and Figure 1.

**Table 1.** Types of cracks in road pavements.

| Flexible Pavements | | Rigid Pavements | |
|---|---|---|---|
| **Distress** | **Cause** | **Distress** | **Cause** |
| Alligator Cracking | load | Corner Break | load |
| Block Cracking Slippage Cracking | traffic | Shattered Slab/Intersecting Cracks | load |
| Longitudinal Cracking | climate | Durability ("D") Cracking | climate |
| Transverse Cracking | climate | Longitudinal, Transverse, and Diagonal Cracking | load |
| Joint Reflection Cracking | climate | Shrinkage Cracks | climate |

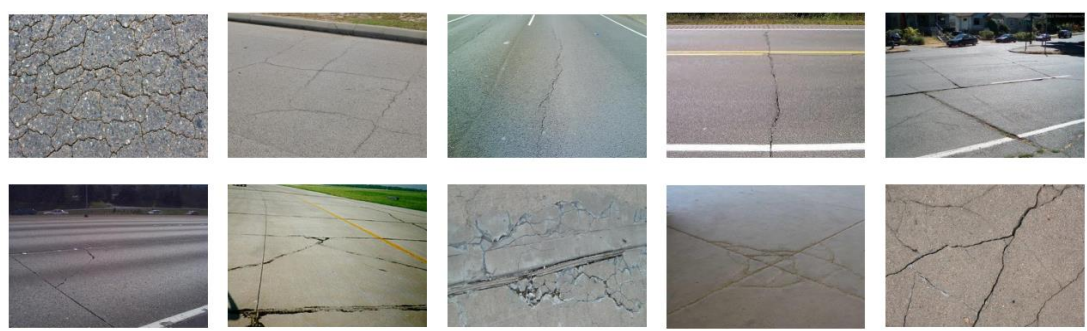

**Figure 1.** Some different crack types are shown. In the top row (from the left to right: alligator cracking, block cracking, slippage cracking, longitudinal cracking, transverse cracking, and joint reflection cracking); on the bottom row (from the left to right: corner break, shattered slab/intersecting cracks, durability ("D") cracking, longitudinal, transverse, and diagonal cracking, and shrinkage cracks).

The cracks can shorten the service life of roads; indeed, the water that can penetrate them can reduce the compaction of the materials of the deeper layers of the pavement with the obvious consequence of a decrease in the load-bearing capacity of the whole structure. In addition, this fact increases the unevenness of the road surface that and is potential threat to road safety [5–11]. Therefore, it is clear that to maintain the pavement in good condition, crack detection is a significant step for pavement management. That step can be performed by both visual inspection and automatic survey. Both methods present good results in terms of distresses analysis, but the automatic crack detection system is more efficient, quick, lower costing than traditional human vision detection. Therefore, automatic crack detection has attracted much attention of scientific and technical corporations in recent years.

*1.2. Monitoring System*

In the past few decades, many researchers have performed structure health monitoring [12–17]. Yu et al. in [18] proposed an integrated system based on the robot for crack detection, which includes mobile manipulate and crack detection system. The mobile manipulate system is used to ensure distance from the objects, and crack detection system is employed to obtain pavement crack information. Oh et al. in [19] proposed bridge detection system, including a designed car, robot system, and machine vision system. Lim et al. in [20] designed a crack inspection system, which consists of three parts: mobile robot, vision system, and algorithm. The camera is mounted on the mobile robot to collect crack images; Laplacian of Gaussian algorithm is applied to extract crack information.

Li et al. in [21] used the laser-image techniques to construct the road surface 3D point clouds. The collecting laser point cloud images are divided into small patches, which is used to identify as containing cracks or not. The minimum spanning tree is employed to extract the cracks from the image patches. Zou et al. in [22] proposed path voting techniques to perform crack detection based on laser range images. Firstly, the local grouping is employed with path voting algorithm based on 3D point cloud images. Then, crack seeds are used for graph representation to extract cracks information. Fernandes et al. proposed a crack detection system by using a light field imaging sensor (Lytro Illum camera), which is employed to disparity information to obtain cracks on the road [23].

*1.3. Crack Detection Algorithms*

Existing visual-based crack detection algorithms can be roughly divided into two branches: traditional crack detection methods and artificial intelligence.

1.3.1. Traditional Crack Detection Methods

- Wavelet transform: Zhou et al. in [24] used a wavelet transform to perform crack detection. Different frequency sub-bands are employed to distinguish crack from images, and high and low amplitudes are defined as crack and noises, respectively. A 2-D wavelet transformation to separate crack and no-crack regions was proposed by Subirats et al. in [25].
- Image thresholding: A threshold value is applied in some research [26–28] to segment crack regions, followed by morphological technologies for refining the processed crack images. The method in [26] needs to preprocess the images with morphological filter to reduce pixels intensity variance, followed dynamic thresholding to detect the cracks. These methods have low efficiency. Oliveira in [26,29] proposed the threshold-based segmentation method. In CrackIT [30], the threshold-based segmentation is proposed to distinguish crack block from the image. After that, they updated their works to CrackIT toolbox [29]. And the latest improvement in [31] used the connectivity consideration as a post-processing step, which contains two steps: selection of prominent "crack seeds" and binary pixels classification, which can improve segmentation results.
- Hand crafted feature and classification: The hand crafted features descriptors are applied to extract crack information from images, followed by patch classifier. [32–34]. Quintana et al. in [34] proposed a computer vision algorithm contains three parts: hard shoulder detection, proposal regions, and crack classification. The Hough transform (HT) was used to detect the hard shoulder; the Hough transform features (HTF) and local binary pattern (LBP) was employed in the proposal regions step; finally, classification was used to detect the crack. It is clear that crack detection operation has low efficiency, and it cannot perform automatic crack detection.
- Edge detection-based methods: Other authors applied the Canny [35] and Sobel [36] edge detector to extract cracks information. Maode et al. in [37] used a modified median filter to remove cracks' noises and the morphological filters were adopted to detect cracks.
- Minimal path-based methods: All these algorithms take brightness and connectivity into consideration for crack detection. Kaul et al. in [38] used the minimal path selection (MPS) method, which is based on fast-marching algorithm to find open and closed curves, and did not employ prior knowledge for endpoints and topology. In addition, the proposed method is fairly robust to the addition of noise. Baltazart et al. proposed three different ongoing improvement with MPS, including selecting crack endpoints, path finding strategy and selection of minimum path cost, and the proposed method can improve the MPS performance in both segmentation and computation time [39]. Nguyen et al. in [40] took brightness and connectivity into consideration for crack detection simultaneously with free-form anisotropy (FFA). In [41], Amhaz et al. introduced the labelled MPS for minimal path selection, which relies on the localization of minimal path based on Dijkstra's algorithm or A* family, and the proposed method can provide robust and precise results. By contrast, Kass et al. in [42] used the theory of actives contours ("snakes"), which used L2 norm for constrained minimization.

1.3.2. Artificial Intelligence

Wang et al. in [43] proposed a multi-class classification method, which applied support vector and machine (SVM) and data fusion to inspect aircraft skin crack. Shi et al. proposed a CrackForest method to describe the crack feature with random structured forests, and the proposed the public CFD database with road crack images was very popular for scholars and researchers [44]. However, these methods are excessive relying on feature descriptors, which is difficult for human to detect different types of crack images.

Recently, with the development of machine learning classified as deep learning inspired by structure of the brain called artificial neural networks (ANN) [45], many algorithms have been proposed to perform object detection and image classification tasks. ANN is employed to solve many civil engineering problems [46–50]. Gao and Mosalam in [51] applied the transfer learning to detect damage images with structural method, and this method can reduce the computational cost by using the pre-trained neural network model. Meanwhile, the author needs to fine the neural network to perform the crack detection. Local patch information was employed to inspect crack information by convolutional neural networks (CNN) in [52]. In CrackNet [53], the algorithm improved pixel-perfect accuracy based on CNN by discarding pooling layers. In CrackNet-R [54], a recurrent neural network (RNN) is deployed to perform automatic crack detection on asphalt road. Cha et al. [55] adopted a sliding windows based on CNN to scan and detect road crack. Fan et al. in [56] proposed a structured prediction method to detect crack pixels with CNN. The small structured pixel images (27 × 27 pixels) was input into the neural network, which may generate overload for the computer memory. Ensemble network is proposed to perform crack detection and measure pavement cracks generated in road pavement [57]. Maeda et al. on [58] adopted object detection network architecture to detect crack images, and the network architecture can be transferred to a smartphone to perform road crack detection. Cha et al. used the Faster-RCNN to inspect road cracks [59]. Yang et al. in [60] adopted a fully convolutional network (FCN) to inspect road pavement cracks at pixel level, which can perform crack detection by end-to-end training. Li et al. in [61] employed the you-only-look-once v3 (YOLOv3)-Lite method to inspect the aircraft structures, and the depth wise separable convolution and feature pyramid were adopted to design the network architecture and joined the low- and high-resolution for crack detection. Jenkins et al. presented an encoder-decoder architecture to perform road crack detection, and the function of the encoder and decoder layers are used to reduce the size of input image to generate lower level feature maps, and obtain the resolution of the input data with up-sampling, respectively [62]. Tisuchiya et al. proposed a data augmentation method based on YOLOv3 to perform crack detection, which can increase the accuracy effectively [63].

It is clear that the feature maps become more and more coarse after several convolution and pooling operations in the CNN process. At the same time, the detailed and abstracted features are presented in large-scale and small-scale layers. Liu et al. in [64] proposed an algorithm to fuse different scale features to improve object detection performance. In the image segmentation process, U-net is proposed in [65] to perform semantic image segmentation based on encoder-decoder architecture to improve accuracy. The dilated convolution for multiple rates is proposed in [66–68] to increase context and obtain more deeper features to improve network performance.

*1.4. Contribution*

Inspired by above observations, in this paper a new network called U-HDN, to fuse multi-scale features in encoder-decoder network based on U-net for crack detection is proposed. The flowchart and the proposed U-HDN architecture are shown in Figure 2 and Figure 3, and the proposed method consists of three components: U-net architecture, multi-dilation module (MDM), and hierarchical feature (HF) learning module. Firstly, an U-net is divided into encoder and decoder networks, which have the same scale at each stage. The encoder networks are applied to extracted features of cracks after convolutions and pooling layers. The decoder networks are employed to restore the image size after a series of up-sampling and convolution layers.

Then, a multi-dilation module (MDM) is designed, which is embedded into an encoder-decoder architecture to obtain cracks features of multiple context sizes. The crack features of multiple context size can be integrated into multi-dilation module by dilation convolution with different dilation rates, which can obtain much more cracks information.

Next, hierarchical feature (HF) learning module is designed to obtain multi-scale feature from the high- to low- level convolutional layers. The single-scale features of each convolutional stage are used to predict pixel-wise crack detection at side output.

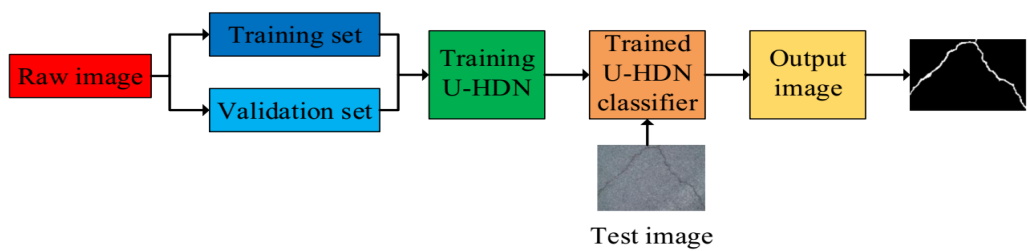

**Figure 2.** Flowchart for detecting pavement cracks.

Finally, the single-scale feature at each side output is concatenated to produce a final fused feature map. Both side outputs and fused results are supervised by deeply-supervised nets (DSN) [69].

The contributions of U-HDN are the following:
1. A new automatic road crack detection method, called U-HDN based on U-net is designed, and encoder-decoder networks are introduced to perform end-to-end training for crack detection. The hierarchical features of crack can be learning in multiple scales and scenes effectively.
2. U-net architecture is modified. Firstly, the pool4, conv9, conv10, and up-conv1 based on U-net model are removed. Secondly, in order to implement end-to-end training, zero-padding during each convolution and up-convolution process are performed.
3. The MDM is proposed to learn crack features of multiple context sizes. The crack features of multiple context size can be integrated into MDM by dilation convolution with different dilation rates.
4. HF learning module is designed to obtain multi-scale feature from the high convolutional layers to low-level convolutional layers. The fusion of hierarchical convolutional features shows a better performance for inferring cracks information.

The rest of this paper is organized as follows: the details of the proposed U-HDN is described in the Section 2 (Methods). Some comprehensive experiments to show the performance for U-HDN and make a comparison with state-of-art algorithms were conducted and the results are discussed in the Section 3 (Experiments and Results). Finally, Section 4 reports the conclusions of the research and some possible future improvements of the method are proposed.

## 2. Methods

In this section, the details of proposed method are introduced, which are the core component of U-HDN. End-to-end classification approach based on encoder-encoder network is employed to perform road crack detection.

The image features are auto-selection in the convolutional operation process, and the selection image features are based on image pixels information from the point of deep learning. Meanwhile, the feature maps tend to be considered and calculated in the convolutional operation process. Therefore, the proposed method is designed and calculated the number of feature maps. In this paper, we employ spatial domain to calculate the feature maps, and the number of the feature maps are shown in Figure 3 (shown on the green boxes).

Deep learning tends to learn image features based on convolutional operation without pre-processing (such as, filter, reducing noises, and data augmentation et al.), according to ground truth, regression function and other active functions. This operation can present wider generalization ability in the database, which can accomplish automatic object detection or semantic segmentation with end-to-end training. Meanwhile, the neural network will auto-learn and extract crack features by convolutional operation, according to the parameters setting and ground truth.

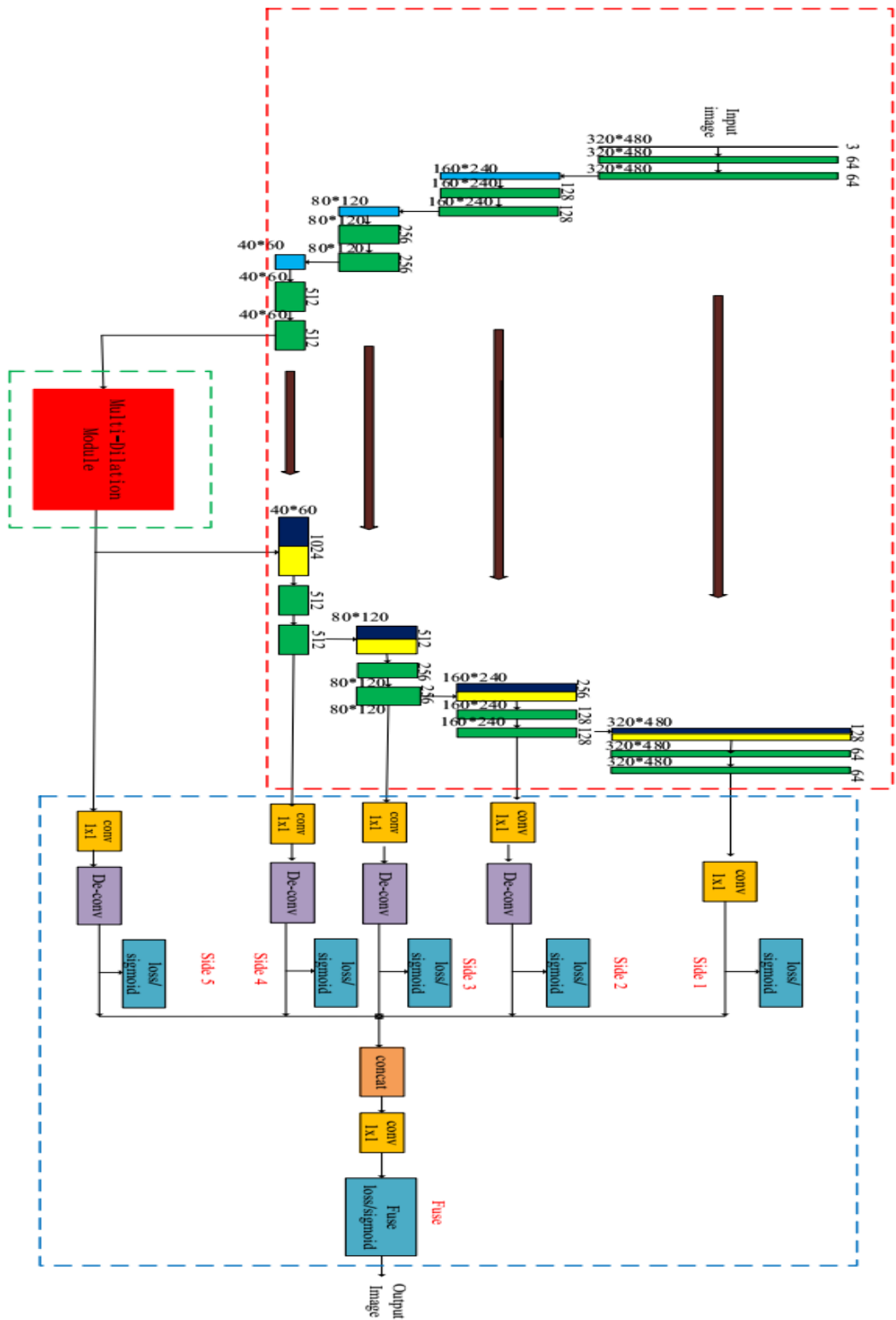

**Figure 3.** The proposed U-HDN architecture consists of three components: U-net architecture, multi-dilation module, and hierarchical feature learning module. The red dotted box presents the modified U-net; the green dotted box is a multi-dilation module; the blue dotted box shows the hierarchical feature learning module.

*2.1. U-Net Architecture*

In this paper, the main backbone of the U-HDN is based on U-net architecture, which is divided into two parts: contracting path (or encoder) and expansive path (or decoder) locating in the left and right side, respectively [65].

As is shown in Figure 3, the red dotted box presents the modified U-net. Contracting path consists of two 3 × 3 convolution layers, each followed by the activation function rectified linear unit (ReLU) [70], and a 2 × 2 max pooling layers for down-sampling.

The expansive path consists of a 2 × 2 up-convolution being up-sampled features, cropped features from the contracting path, and two 3 × 3 convolution layers, each followed by the activation function ReLU. In this U-net architecture, the components pool4, conv9, conv10, and upconv1 were removed. Secondly, in order to implement end-to-end training, a transformation zero-padding during each convolution and up-convolution process was performed. Meanwhile, in order to understand the convolution neural network, we recommend readers to look up this article [71].

Convolution layer: $k$ filters (or kernels) belong to the convolutional layer with the weight $w$. In the convolution process, input image being convolving with filters and plus bias $b$ that can obtain $k$ feature maps. In order to increase nonlinearity for output, ReLU is employed as activation function after convolution process.

Max pooling layer: max pooling is applied to obtain maximum value for each subarray during down-sampling process, and this operation can reduce computational complexity.

Activation Function: the activation function ReLU to increase nonlinearity for convolution layes' output was used. At the same time, the sigmoid function to distinguish crack and non-crack pixels for final output result was adopted [72]. Zero-padding: it is convenient to pad the input matrix with zeros around the border, so that we can apply the filter to bordering elements of our input image matrix. The function of zero padding can ensure the size of the output image that we desired during the up-sampling process [65,73].

*2.2. Multi-Dilation Module (MDM)*

In encoder network of U-net, only one type of the convolutional filters is employed to obtain receptive field for extracting crack features, which has a negative influence on detecting different cracks types, such as, vertical, horizontal and topologies.

Therefore, a MDM based on encoder features to obtain multiple context sizes' features was designed [66–68], as is shown in Figure 4. The dilation convolution is able to expand the sizes of the convolution filters, instead of using larger filter and down-sampling. The MDM has a better performance for extracting and detecting cracks with multiple context sizes.

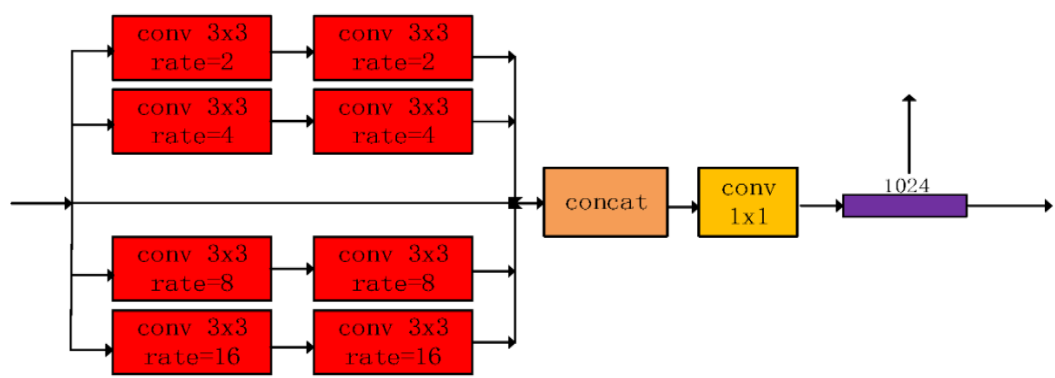

**Figure 4.** The overview of the multi-dilation module.

In a 2-D signal, the dilation convolution is defined as the following equation [66]:

$$y[i] = \sum_{k=1}^{K} x[i + r \cdot k] w[k] \qquad (1)$$

where $x[i]$ and $y[i]$ are input and output signal for each location $i$, respectively. $w[k]$ is defined as the filter of length $K$. Dilation rate $r$ corresponds to stride for sampling input signal. It is necessary to insert a number of $r-1$ zeros between two consecutive filter values along each spatial dimension in the process of convolution operation. In the standard convolution operation, it can be assumed $r=1$.

Assuming a convolution filter size equal to $k$, the dilation convolution filter size is $k_d$ [66].

$$k_d = k + (k-1) \times (r-1) \tag{2}$$

As is shown in Figure 5, different dilation rates are designed for convolution filters. Although the dilation convolution expands feature context size in the convolution, it does not increase amount of calculation with inserting of $r-1$ zeros.

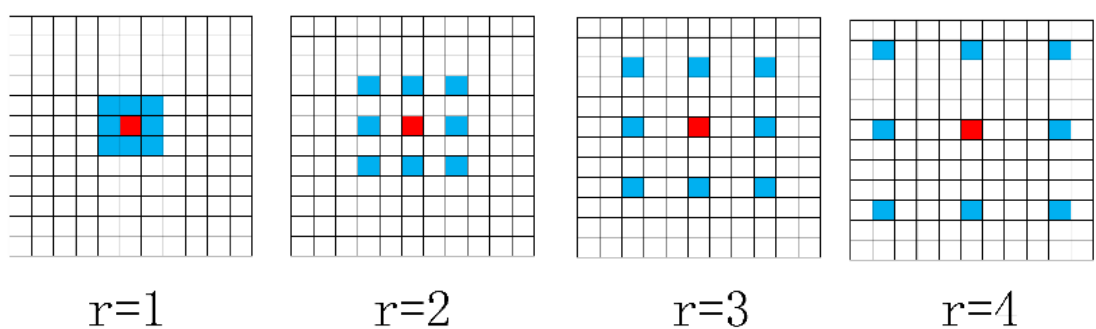

**Figure 5.** Convolution filters with different dilation rates.

Due to complex road images, different topologies and width, standard convolution can only obtain one context, which cannot effectively satisfy both thin, simple cracks and wide, complex cracks.

Therefore, a multi-dilation module (MDM) to address above problems was proposed. This module uses the different context sizes for crack features and fuses them to get multiple context features. Firstly, the four dilation rates are defined as 2, 4, 8, and 16, respectively. These four dilation convolution operations are able to extract crack features with different context sizes. After that, the five different crack features by a concatenation method were combined. Next, a 1 × 1 convolution to change the number of features from 512 × 5 to 1024 was used. After this convolution operation, the multi-dilation module was accomplished, to obtain output features that can have a better performance for various crack types.

*2.3. Hierarchical Feature (HF) and Loss Function*

Since the high-level feature maps have more complex context information than low levels during the deeper convolution operation. Therefore, the HF learning network was adopted (or side 1, 2, 3, 4, 5, and fused), which can perform crack detection individually. A real example is shown in Figure 6, it shows the ground truth for input image and the fused feature maps ant different scales.

Each side outputs and fused output are supervised by DSN [69] with holistically-nested edge detection (HED) [74] for edge detection. The HED for crack detection was introduced. A training database is defined as $S=(X_n,Y_n), n=1,\ldots,N$, where $X_n$ and $Y_n$ are the raw input image and ground truth crack map, respectively. In order to write convenience, the subscript $n$ is dropped in subsequent paragraphs. $W$ and $M$ are defined as the number of network parameters and side networks, respectively.

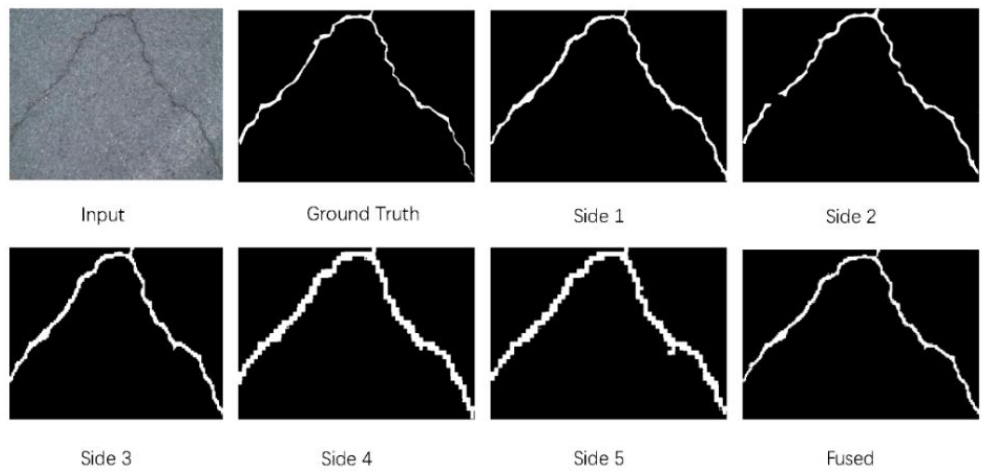

**Figure 6.** A real example of crack detection based on U-HDN. It shows the comparison between ground truth for input image and fused feature maps at different scales.

Each side network is followed by a classifier and the weights for each side network is denoted as $w = (w^{(1)}, \ldots, w^{(M)})$. The following equation is the loss function for side networks [74].

$$L_{side}(W, w) = \sum_{m=1}^{M} \alpha_m l_{side}^m (W, w^m) \qquad (3)$$

where $l_{side}$ is the image-level loss function for each side network. The parameter $\alpha_m$ is a hyperparameter for loss weight at each side-out layer. In this project, $M = 5$. During end-to-end training, the image pixels are divided into crack and non-crack pixels with a classifier. Therefore, crack detection can be denoted as a binary classification problem. An activation function sigmoid is applied to distinguish the non-crack and crack pixels. Furthermore, the sigmoid cross entropy loss function to address imbalance samples problem was modified. This sigmoid loss function [65] with weight is shown Equation (4):

$$l_{side} = \frac{1}{N} \sum_{i=1}^{N} \{\beta \, y_i \log \hat{y}_i + \gamma(1 - y_i)\log(1 - \hat{y}_i)\} \qquad (4)$$

where $\beta$ and $\gamma$ are hyperparameters, $N$ is defined as the pixels' number for one image. $y_i$ and $\hat{y}_i$ are the ground truth and predicted output result locating $i^{th}$ pixel, respectively.

Each the side network can generate a prediction feature map, which consists of a single output loss. The entitle outputs of side network are fused to generate final prediction result with concatenation method, and the fused loss function is equal to $l_{side}$:

$$l_{fuse} = l_{side} \qquad (5)$$

Finally, the total loss function of the entitle network is defined as following equation:

$$L_{total} = L_{side} + l_{fuse} \qquad (6)$$

## 3. Experiments and Results

In this part, the implementation details for the proposed U-HDN are described. Then, evaluation metric and compared methods are presented. Finally, the experimental results are analyzed.

### 3.1. Implementation Details

The proposed U-HDN is programmed by Pytorch library [75] as the deep learning framework for training and testing under Google Colaboratory (free with time limitation) GPU Workstation with the types of Tesla P100-PCIE-16 GB, memory 16280 MB.

The public databases CFD [44] and AigleRN [76] were used to train and test the proposed network, which do not demonstrate the visual condition for image collection. The CFD database contains 118 color images (images of size 320 × 480 pixels), which was collected by iPhone 5 smartphone in Beijing, China. In this project, a sample of 72 images were used to train the method and a sample of 46 images were used to test the proposed U-HDN. The AigleRN database includes 38 gray images (with two types of images' size: 991 × 462 pixels and 311 × 462 pixels), which was obtained from a sample of pavements located in France. At the same time, the 24 images and 14 images were employed to train and test the U-HDN, respectively. In this paper, to extract the crack pixels, and distinguish the crack and non-crack pixels some procedures were performed. The images of both public databases have a resolution equal to 600 ppi; this means that the images were acquired with each pixel corresponding to approximately 1 mm$^2$ of the real road pavement.

The visual condition for these two database was collected at vertical incidence [44]. The results in this research would not have the goal to demonstrate the effect of visual condition, for this reason, this information is not considered important and it was not reported.

At this moment, the proposed method is not able to detect the crack widths, but the calculus of this important characteristic will be obtained in the next upgrade of the model. In this paper, we perform to extract the crack pixels, and distinguish the crack and non-crack pixels.

The training time for *CFD* is about 5 h and 20 min. The training time for AigleRN is about 3 h.

3.1.1. Parameters Setting

The hyperparameters contain: bath size (4 images for CFD, 1 image for AigleRN), optimizer (adam), learning rate (0.001), min-learning rate (0.000001), learning rate scheduler (plateau), patience (10), factor (0.95) with two functions (torch.optim.lr_scheduler.ReduceLROnPlateau and torch.optim.Adam based on Pytorch library [75]). These parameters are intrinsic parameter during training the neural network, such as learning rate. When we train the CFD, 4 images are input the neural network once time; When we train the AigleRN, 1 image is input the neural network once time. This setting can enable crack detection to obtain global optimum in the segmentation performance. We fix the parameters setting for these two databases during training neural network.

3.1.2. Evaluate Metrics

The models considered in this study were evaluated by three performance measures: the precision ($Pr$), the recall ($Re$), and the F1 score ($F1$). The precision and recall [77] are calculated by Equations (7) and (8) as below:

$$Pr = \frac{TP}{TP + FP} \tag{7}$$

$$Re = \frac{TP}{TP + FN} \tag{8}$$

where $TP, FP$, and $FN$ are the number of the true positive, false positive and false negative, respectively. $F1$ is employed to evaluate the overall performance for the crack detection and it is the harmonic average of Precision and Recall [77] calculated by Equation (9).

$$F1 = \frac{2 \times Pr \times Re}{Pr + Re} \tag{9}$$

Specifically, two different metrics based on $F1$ are adopted in the evaluation: the best $F1$ on the public database for a fixed threshold (ODS), and the aggregate $F1$ on the public database for the best threshold in each image (OIS) [78].

The definitions of the ODS and OIS are reported in the Equations (10) and (11):

$$\text{ODS} = max\left\{\frac{2 \times Pr_t \times Re_t}{Pr_t + Re_t} : t = 0.001, 0.002, \dots, 0.999\right\} \tag{10}$$

$$\text{OIS} = \frac{1}{N_{img}} \sum_{i}^{N_{img}} max \frac{2 \times Pr_t^i \times Re_t^i}{Pr_t^i + Re_t^i} : t = 0.001, 0.002, \dots, 0.999 \quad (11)$$

The values $t, i$, and $N_{img}$ are the threshold, index and the number of the images. The parameters $Pr_t, Re_t, Pr_t^i$ and $Re_t^i$ are precision and recall based on threshold $t$ and image $i^{th}$, respectively.

For the proposed U-HDN, the transitional areas between non-crack and crack pixels were considered before computing $TP, FP,$ and $FN$. Considering the subjective manual labels for ground truth, the transitional areas (2 pixels distance) between crack and non-crack pixels are accepted in these papers [41,56,57,79,80]. Therefore, 2 pixels of distance is accepted in this project. The decision threshold is defined as 0.5 to obtain a binary output.

### 3.2. Discussion for Multi-Dilation Module (MDM)

The dilation rate presented in Equation (1) plays an important role in varying the context size based on MDM for the U-HDN. A large dilation rate can obtain a large context size, as is shown in Figures 2 and 3. Specifically, different dilation rates can get different context size, which can produce different prediction results. In order to analyze the different effect of dilation rates, an experiment to proof the setting of the hyperparameters in MDM was performed.

Three groups of $\{1, 2, 3, 4\}, \{1, 2, 4, 8\}, \{2, 4, 8, 16\}$ are tested based on public database CFD and AigleRN. As shown from the experimental results in Table 2 and Table 3, group of $\{2, 4, 8, 16\}$ can obtain the highest accuracy on both databases. The reason is that a large dilation rate can get more context information of the cracks for the relatively wide or thin crack structure, which can improve the crack detection accuracy.

**Table 2.** Experimental results for different dilation rates on CFD database.

| Dilatation Rates | Precision | Recall | F1 Score |
|---|---|---|---|
| $\{1, 2, 3, 4\}$ | 0.943 | 0.933 | 0.935 |
| $\{1, 2, 4, 8\}$ | 0.944 | 0.934 | 0.937 |
| $\{2, 4, 8, 16\}$ | 0.945 | 0.936 | 0.939 |

**Table 3.** Experimental results for different dilation rates on AigleRN database.

| Dilatation Rates | Precision | Recall | F1 Score |
|---|---|---|---|
| $\{1, 2, 3, 4\}$ | 0.914 | 0.921 | 0.915 |
| $\{1, 2, 4, 8\}$ | 0.919 | 0.923 | 0.921 |
| $\{2, 4, 8, 16\}$ | 0.921 | 0.931 | 0.924 |

### 3.3. Experimental Results on CFD

The experimental results of some specimen detection are shown in Figure 7 and Table 4 based on *CFD*. It is clear that Canny and local threshold are sensitive to the noises, which can lead to a negative influence for crack detection.

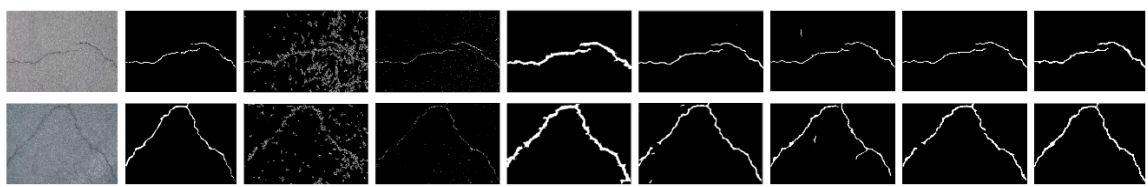

**Figure 7.** Results of comparison of proposed U-HDN with other method based on public database (From left to right: input image, ground truth, Canny, local threshold, CrackForest, structured prediction, U-net, ensemble network, and proposed U-HDN).

Compared with ground truth, it is also observed that CrackForest algorithm can over-measure the number of cracks and extract the wider cracks with a high recall 0.9514, as shown in Table 4. As is shown in Figure 5, although structured prediction and U-net can get a better performance for crack detection, these methods can detect several wrong non-crack pixels. Although ensemble network (threshold = 0.6) can achieve high precision, recall and F1 score, this method can produce resource redundancy and also occur missed detection in the images, as is shown in Figure 7. At the same time, this method cannot perform end-to-end training. The values for two images in Figure 7 are: Pr: 0.978, Re: 0.973, F1: 0.975 (top image) and Pr: 0.977, Re: 0.966, F1: 0.971 (bottom image).

Table 4. Crack detection results on CFD.

| Methods | Tolerance Margin | Pr | Re | F1 |
|---|---|---|---|---|
| Canny [35] | 2 | 0.4377 | 0.7307 | 0.457 |
| Local thresholding [26] | 2 | 0.7727 | 0.8274 | 0.7418 |
| CrackForest [44] | 2 | 0.7466 | 0.9514 | 0.8318 |
| CrackForest [44] | 5 | 0.8228 | 0.8944 | 0.8517 |
| MFCD [81] | 5 | 0.899 | 0.8947 | 0.8804 |
| Method [79] | 2 | 0.907 | 0.846 | 0.87 |
| Structed prediction [56] | 2 | 0.9119 | 0.9481 | 0.9244 |
| Ensemble network (threshold = 0.6) [57] | 2 | 0.9552 | 0.9521 | 0.9533 |
| Ensemble network (threshold = 0.5) [57] | 2 | 0.9256 | 0.9611 | 0.934 |
| U-net [65] | 2 | 0.9325 | 0.932 | 0.928 |
| U-net + HF | 2 | 0.933 | 0.933 | 0.931 |
| U-net + MDM | 2 | 0.9302 | 0.931 | 0.93 |
| U-HDN | 2 | 0.945 | 0.936 | 0.939 |

The proposed U-HDN can perform end-to-end training and also obtain a satisfactory accuracy than other algorithms (Pr: 0.945, Re: 0.936, F1: 0.939). The main reason is that U-HDN can extract and fuse different context sizes (based on MDM) and different levels (high-level, and low-level based on HF) feature maps than other algorithms. In Table 5, it is clear that proposed U-HDN achieves superior performance compared to other algorithms in terms of ODS and OIS.

Table 5. The ODS, and OIS of comparison methods on CFD.

| Methods | ODS | OIS |
|---|---|---|
| HED [74] | 0.593 | 0.626 |
| RCF [64] | 0.542 | 0.607 |
| FCN [82] | 0.585 | 0.609 |
| CrackForest [44] | 0.104 | 0.104 |
| FPHBN [78] | 0.683 | 0.705 |
| U-net [65] | 0.901 | 0.897 |
| U-HDN | 0.935 | 0.928 |

*3.4. Experimental Results on AigleRN*

The experimental results of some specimen detection are shown in Figure 8 and Table 6 based on AigleRN database include 38 images. As shown in Figure 6, it is observed that two traditional methods (Canny and local threshold) cannot extract the crack skeleton and detect the continuous cracks, which are susceptible to the noises. It is clear that FFA and MPS are able to inspect local and small cracks but also fail to extract crack skeleton and find continuous cracks. Although the structured predicted method can extract rough skeleton and detect cracks, it can also occur missed detection in the images. The ensemble network is able to obtain a better crack skeleton than structured predicted, but it cannot find cracks that are more continuous. The values for two images in Figure 7 are: Pr: 0.915, Re: 0.961, F1: 0.937 (top image) and Pr: 0.924, Re: 0.981, F1: 0.952 (bottom image).

Meanwhile, it is clear that FFA can detect thicker crack than our proposed, and cannot extract the crack skeleton, which can cause the low precision rate, as is shown in Table 6. The method proposed is able to extract the crack skeleton. Secondly, it is observed that the method can obtain much more number of false positive than false negative, which lead to the higher recall rate than precision rate. Then, the 2-pixel distance can also help to improve the precision rate. Finally, the average vales based on test database can improve the global precision rate.

The proposed U-HDN method can achieve superior performance compared to other algorithms, as is shown in Figure 6 and Table 6 (Pr: 0.921, Re: 0.931, F1: 0.924). The main reason is that U-HDN can extract and fuse different context sizes (based on MDM) and different levels (high-level, and low-level based on HF) feature maps than other algorithms. Hence, U-HDN can get a high accuracy.

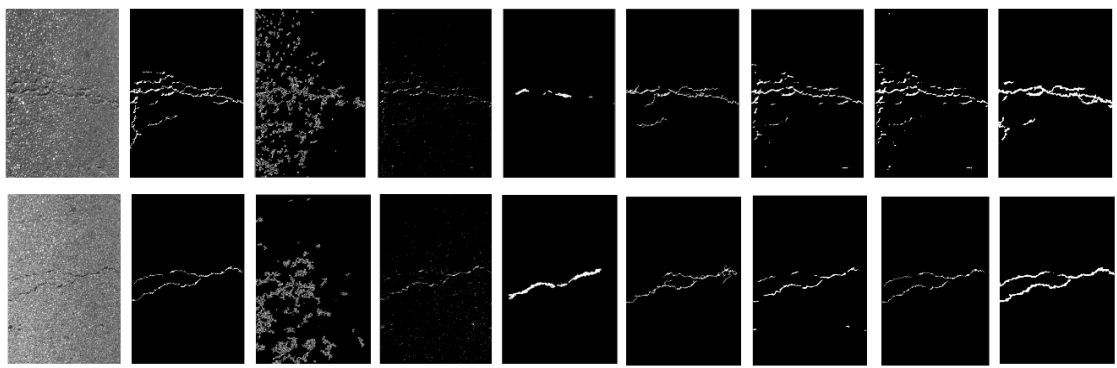

**Figure 8.** Results of comparison of proposed U-HDN with other method based on public database (From left to right: input image, ground truth, Canny, local threshold, FFA, MPS, structured prediction, ensemble network, and proposed U-HDN).

### 3.5. AigleRN Dataset Generalization

As reported above, the *AigleRN* database include 38 images (two types of resolution: 991 × 462 and 311 × 462). ESAR database (resolution 768 × 512) is collected by a statistic system, which contains 15 images. LCMS database includes 5 images. Because of having small number of images for these databases, they are combined to obtain a new database, named AEL with in total 38 + 15 + 5 = 58 images. In Table 7, it is clear that proposed U-HDN achieves high performance compared with other algorithms in terms of ODS and OIS.

**Table 6.** Crack detection results on AigleRN.

| Methods | Tolerance Margin | Pr | Re | F1 |
|---|---|---|---|---|
| Canny [35] | 2 | 0.1989 | 0.6753 | 0.2881 |
| Local thresholding [26] | 2 | 0.5329 | 0.9345 | 0.667 |
| FFA [43] 12 | 2 | 0.7688 | 0.6812 | 0.6817 |
| MPS [42] | 2 | 0.8263 | 0.841 | 0.8195 |
| CrackForest [44] | 2 | 0.8424 | 0.801 | 0.8233 |
| CrackForest [44] | 5 | 0.9028 | 0.8658 | 0.8839 |
| Structed prediction [40] | 2 | 0.9178 | 0.8812 | 0.8954 |
| Method [67] | 2 | 0.869 | 0.9304 | 0.8986 |
| Ensemble network (threshold = 0.6) [57] | 2 | 0.9302 | 0.9266 | 0.9238 |
| Ensemble network (threshold = 0.5) [57] | 2 | 0.9334 | 0.8879 | 0.9211 |
| U-net [65] | 2 | 0.9127 | 0.9076 | 0.91 |
| U-net + HF | 2 | 0.911 | 0.922 | 0.913 |
| U-net + MDM | 2 | 0.9138 | 0.9245 | 0.914 |
| U-HDN | 2 | 0.921 | 0.931 | 0.924 |

**Table 7.** The ODS, and OIS of comparison methods on AEL.

| Methods | ODS | OIS |
|---|---|---|
| HED [74] | 0.042 | 0.626 |
| RCF [64] | 0.462 | 0.607 |
| FCN [82] | 0.322 | 0.609 |
| CrackForest [44] | 0.231 | 0.104 |
| FPHBN [78] | 0.492 | 0.705 |
| U-net [65] | 0.752 | 0.897 |
| U-HDN | 0.783 | 0.928 |
| U-HDN (only using AigleRN) | 0.927 | 0.912 |

## 4. Conclusions

The analysis and survey of pavement crack plays an important role in the road and airport pavement management system. In this project, the proposed U-HDN method can achieve a high precision and accuracy for pavement crack detection. An MDM and HF module based on U-net are developed in this paper. The MDM is able to obtain and extract feature maps of different context sizes by different dilation rates. The HF module can obtain multi-scale (high-level and low-level) feature maps, which can be integrated to predict pixel-wise crack detection at side output. By combining two MDM and HF in the U-net, U-HDN can achieve a satisfactory performance.

Although the proposed U-HDN can obtain a satisfactory performance than other methods, the neural network is a complicated structure which contains redundant feature maps and cause computational cost and low efficiency. These issues will be addressed in the future work.

- In order to remove the redundant features maps, the channel pruning and automatically designing neural network will be explored to improve the computational efficiency and accuracy.
- Some methods tend to research crack detection for static images. Actually, video streaming detection also has a significant function for road cracks. Therefore, we will study this direction in the future work.
- We plan to propose a new method to address the cement concrete crack detection, evaluate the global surface waterproofing and repair water-leakage cracks.
- Due to F1 sensitivity to the pixel margin, it is not appropriate for author to compare the performance segmentation algorithms that do not give all the details on the metric. Therefore, we will try contact some authors to obtain the source codes and analyze them, followed by exploring and constructing an integrated crack detection system.

**Author Contributions:** Conceptualization, Z.F.; methodology, C.L.; software, Y.C., X.C., and J.W.; validation, X.C.; formal analysis, C.L.; investigation, Y.C. and J.W.; resources, C.L.; data curation, C.L.; writing-original draft preparation, C.L.; writing-review and editing, Z.F., P.D.M. and G.L.; visualization, C.L.; supervision, Z.F., P.D.M. and G.L.; project administration, Z.F., G.L. All authors have read and agreed to the published version of the manuscript.

**Funding:** This work was supported by the Science and Technology Planning Project of Guangdong Province of China under grant 180917144960530, by the Project of Educational Commission of Guangdong Province of China under grant 2017KZDXM032, by the State Key Lab of Digital Manufacturing Equipment and Technology under grant DMETKF2019020, by the Project of Robot Automatic Design Platform combining Multi-Objective Evolutionary Computation and Deep Neural Network under grant 2019A050519008, and by the China Scholarship Council (CSC) in 2019.

**Conflicts of Interest:** The authors declare no conflict of interest.